\documentclass[11pt]{article}

\usepackage[final]{acl}
\usepackage{times}
\usepackage{latexsym}

\usepackage[T1]{fontenc}
\usepackage[utf8]{inputenc}

\usepackage{microtype}

\usepackage{inconsolata}

\usepackage{graphicx}

\usepackage{hyperref}
\usepackage{graphicx}
\usepackage{multirow}
\usepackage{hwemoji}
\usepackage{fancyvrb}
\usepackage{natbib}
\usepackage{amssymb}
\usepackage{tcolorbox}
\tcbuselibrary{skins,breakable}

\usepackage{enumitem}
\setitemize{noitemsep,itemsep=3pt,topsep=3pt}
\setenumerate{noitemsep,itemsep=3pt,topsep=3pt}

\begin{document}

\title{Evaluating Human-LLM Representation Alignment: A Case Study on Affective Sentence Generation for Augmentative and Alternative Communication}

\author{ \textbf{Shadab Choudhury\textsuperscript{1}},
 \textbf{Asha Kumar\textsuperscript{2}},
 \textbf{Lara J. Martin\textsuperscript{1}}
\\
 \textsuperscript{1}Computer Science and Electrical Engineering Department\\
 \textsuperscript{2}Information Systems Department\\
 University of Maryland, Baltimore County\\
  \href{mailto:shadabc1@umbc.edu,laramar@umbc.edu}{\texttt{\{shadabc1,laramar\}@umbc.edu}}\\
}
\maketitle
\begin{abstract}
Gaps arise between a language model's use of concepts and people's expectations. This gap is critical when LLMs generate text to help people communicate via Augmentative and Alternative Communication (AAC) tools. In this work, we introduce the evaluation task of Representation Alignment for measuring this gap via human judgment. In our study, we expand keywords and emotion representations into full sentences. We select four emotion representations: Words, Valence-Arousal-Dominance (VAD) dimensions expressed in both Lexical and Numeric forms, and Emojis.  In addition to Representation Alignment, we also measure people's judgments of the accuracy and realism of the generated sentences. While representations like VAD break emotions into easy-to-compute components, our findings show that people agree more with how LLMs generate when conditioned on English words (e.g., ``angry'') rather than VAD scales. This difference is especially visible when comparing Numeric VAD to words. Furthermore, we found that the perception of how much a generated sentence conveys an emotion is dependent on both the representation type and which emotion it is.
\end{abstract}

\section{Introduction}
%%%%%%%%%%%%%
Augmentative and Alternative Communication (AAC) tools help people who cannot communicate verbally to hold conversations. 
Speed of communication is one of the biggest pain points that users point out about their AAC tools \cite{Trnka_2007}. As a result of this lag, users are often talked over, ignored, or otherwise disrespected in conversations \cite{kane_at_2017}.
Fortunately, due to their medium, high-tech AAC options such as phone or tablet applications present the opportunity to incorporate NLP techniques to improve communication speed.

To work toward speeding up AAC use, keyword-based sentence generation and word prediction have been explored in AAC technologies for decades \cite{Trnka_2007,Wiegand_2012,Garcia-etal-2014,fontana-de-vargas-moffatt-2021-automated,shen_kwickchat_2022}.
Unsurprisingly, we have also seen a recent uptick in the use of LLMs in AAC tools \cite{valencia_less_2023,Kitayama_2024,Francis_2024,Bailey_2024}.

\begin{figure}[t!]
    \centering
    \includegraphics[width=\linewidth]{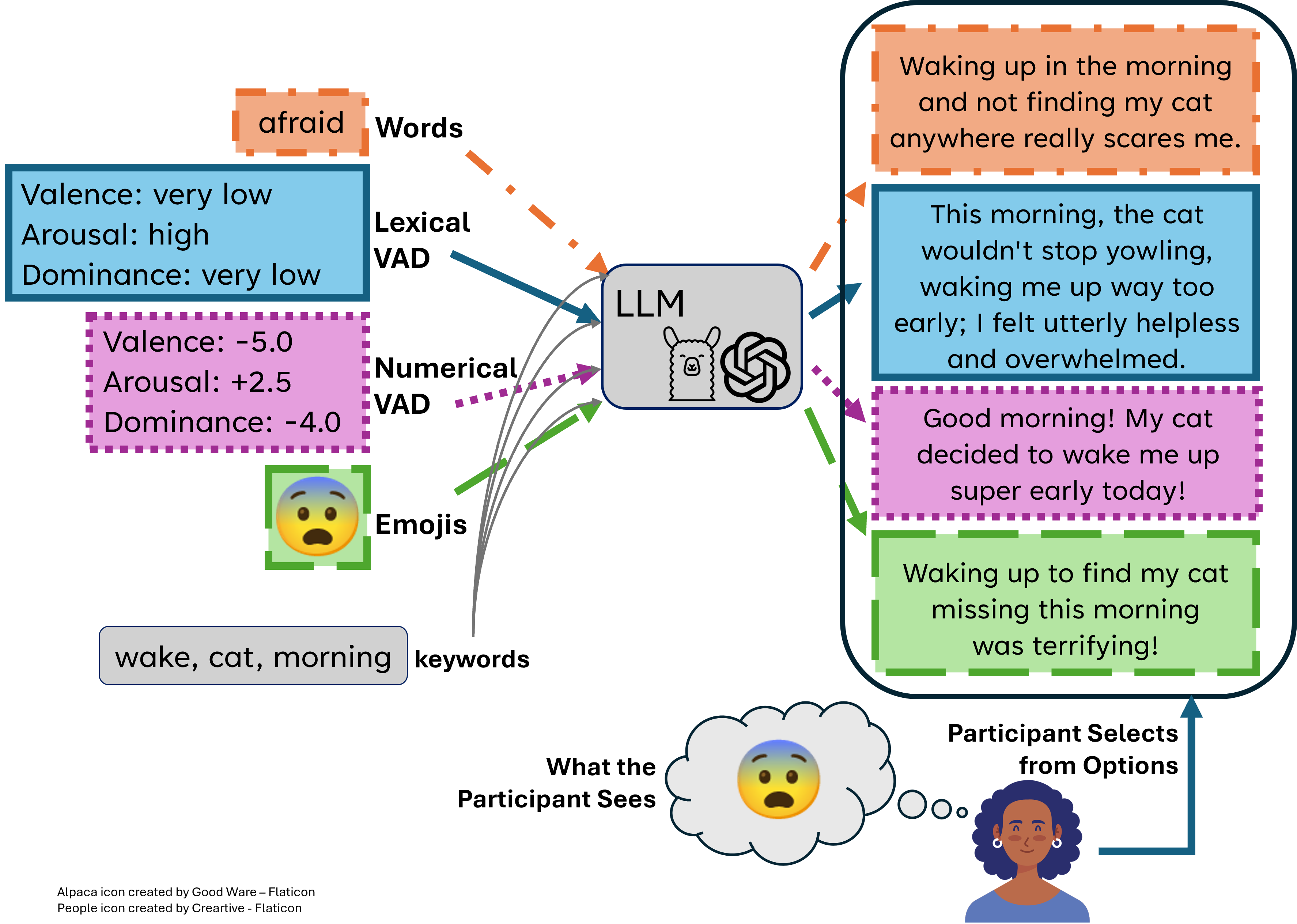}
    \caption{Representation Alignment experiment. Three keywords and an emotion from one of the four representations are used to generate a sentence. Participants are shown the emotion in only one of the representations and select the sentence that best fits that emotion.}
    \label{fig:header}
\end{figure}

There have been fairly successful attempts for personalized generation \cite{Zhang_2023}, such as through prompt engineering, keyword-based generation \cite{hokamp_lexically_2017,Yao_2019} or style transfer \cite{Liu_2024_Customizing}. 
While LLMs can generate human-like text better than their predecessors, there are concerns about putting them into AAC applications. Namely, how much control users have over the LLM, whether the LLM accurately captures the user's way of speaking, or if the LLM is generating text that is appropriate to the conversation's context \cite{valencia_less_2023,martin_bridging_2024}. 
Part of this context includes the current emotional state of the AAC user. 
As such, we wanted to see how well LLMs and humans aligned in their understanding of various emotion representations. Our hope is that representations used in future AAC applications will generate sentences that match the user's intended emotion. 

In this paper, we introduce the task of \textsc{Representation Alignment}, illustrated in Figure \ref{fig:header}, where the alignment between the LLM's output and the human's mental model of a concept is measured. Note that this is different from Theory of Mind, as the LLM is not trying to ascertain the current mental state of the user. Representation Alignment looks at how well an LLM's use of concepts aligns with human understanding. Specifically, we will be focusing on the Representation Alignment of \textit{emotions}.

There are multiple ways an emotion could be represented in the input. One representation could be easier for an LLM to parse, but a different representation might be more accessible to users. Hence, to assess emotion representations in the context of keyword-based sentence generation, we investigate the following research questions:

\begin{enumerate}
    \item Do LLMs' use of emotion representations match humans' expectations?
    \item Is there a preferred representation for conveying emotions when performing keyword-based sentence generation?
\end{enumerate}

We look at four ways of representing emotions: with Words (the English word for a particular emotion), via emojis -- which have become an effective way to express emotions visually over text \cite{kaye_are_2023}, and two types of Valence-Arousal-Dominance scales. We will explain the representation implementation in Section \ref{sec:sentence-generation}.

Valence-Arousal-Dominance (VAD) \cite{mehrabian_basic_1980} scales measure emotions on three axes: Valence (or pleasure) indicating whether its a positive or negative feeling, Arousal indicating how much energy is behind the feeling, and Dominance indicating how much control the user has over that feeling. Although VAD has origins in psychology research, it has been useful for NLP-related areas such as affective computing \cite{mohammad_obtaining_2018, guo_enhancing_2021,el-haj-takanami-2023-unifying} and social computing \cite{Hutto_2014,khosla-2018-emotionx,garg2023mental}. Thus, we were curious about the effectiveness VAD for text generation.

In this work, we addressed the research questions by generating sentences using each of the four emotion representations by few-shot prompting two large language models (GPT-4 and LLaMA-3). We ran a human participant study to determine how well the representations aligned to participants' expectations and how accurately \& realistically LLMs can generate sentences using a particular representation. Our contributions are as follows:

\begin{enumerate}
    \item We introduce a human evaluation paradigm for measuring the alignment between mental models of concepts (such as emotions) and how they are used by LLMs.
    \item We show that humans and LLMs align best when Words, or to a lesser extent, Lexical VAD are used to represent emotions.
    \item We also show that either Words or Lexical VAD, when used in a prompt, give realistic sentences that most participants agree sound like something they would say.
\end{enumerate}

%%%%%%%%%%

\section{Related Work}

\subsection{Keyword-based Sentence Generation}
\label{sec:generation}
Keyword-based sentence generation, also referred to as \textit{lexically-constrained generation}\footnote{We use the two terms interchangeably.}, is a subtopic of controlled text generation where the input to a model is a set of keywords and the output is a sentence that uses those keywords. Early on, \citet{kasper_flexible_1989, uchimoto_text_2002} used grammar systems to build sentences using keywords as anchors. More recently, \citet{mou_sequence_2016, he_show_2021} used sequential models and revised the output repeatedly at each sampling step. \citet{Yao_2019, Wang_2020, Ammanabrolu_2020} similarly generated sequential outputs by prompting with a list of keywords. With transformer-based language models, it's also possible to use the decoder or special tokens to control keywords, style or sentiment \citep{kumar_controlled_2021, samanta_fine-grained_2020, nie_lexical_2022, Krause_2021, Sasazawa_2023}.

However, most of these works focus on smaller language models. \citet{chen_evaluating_2024} showed that LLMs have strong baseline lexically-constrained generation ability. To our knowledge, not much other work has been done in this area using LLMs.

\subsection{Emotion-conditioned Sentence Generation}

Emotional sentence generation has been widely studied as part of style-transfer or empathetic dialogue generation problems. Early work on emotional sentence generation, such as \citet{polzin_emotion-sensitive_2000}, relied on rule-based systems. \citet{ghosh_affect-lm_2017, song_generating_2019} both conditioned the output of a recurrent neural network on specific emotional words, while \citet{singh_adapting_2020} also conditioned on words but used GPT-2 \cite{radford_language_nodate} as the base model. \citet{colombo_affect-driven_2019, lubis_eliciting_2018} utilized the VAD space, adding an emotional vector to the internal representation of the text. \citet{zhou_mojitalk_2018} conditioned a variational autoencoder on Emojis instead.

LLMs have also been used for emotional text generation. A variety of methods like chain-of-thought, retrieval-augmented generation, prompt tuning, etc. have all been successfully used \cite{li_enhancing_2024, mishra_real-time_2023, rasool_emotion-aware_2024, yang_enhancing_2024, resendiz_llm-based_2025, zhang_affective_2024}. However none of these works deal with non-emotional constraints, nor cover the use of VAD scales.

\subsection{Value Alignment}

There have been multiple methods for integrating values into AI such as learning normative behavior from children's stories \cite{AlNahian_2020}, integrating logic \cite{Kim_2021}, using
actor-critic models\cite{Liu_2022}, integrating situated annotations in reinforcement learning from human feedback \cite{Arzberger_2024}, or matching behavior to underlying ethics rather than human actions \cite{Rigley_2025}. 
By finding and comparing explicit representations for ethical values, these methods could potentially be evaluated using our Representation Alignment evaluation technique.
There have been recent efforts to measure the value alignment of existing models \cite{Norhashim_2024}, but only for a binary concept (moral or not moral).

\section{Sentence Generation}
\label{sec:sentence-generation}

\begin{table*}[t]
    \centering
\small
    \begin{tabular}{l|c|c|l|l|l}
        \textbf{Category} & \textbf{Words} &  \textbf{Emoji} &\textbf{Valence} & \textbf{Arousal} & \textbf{Dominance}\\
        \hline
        \multirow{3}{*}{Happy} & Grateful & {\LARGE 😄}& Very High (+2.5)& Moderate (0.0)& Low (-2.5)\\
        & Joyful &{\LARGE 😂}& Very High (+4.0)& High (+1.0)& High (+1.0)\\
        & Content &{\LARGE 😌}& Very High (+4.0)& Moderate (0.0)& Very High (+4.0)\\
        \hline
        \multirow{2}{*}{Surprise} & Surprised &{\LARGE 😯}& High (+1.0)& Very High (+2.5)& Low (-2.5)\\
        & Excited &{\LARGE 😃}& Very High (+2.5)& Very High (+4.0)& High (+1.0)\\
        \hline
        \multirow{2}{*}{Pride} & Impressed &{\LARGE 🙂}& High (+1.0)& High (+1.0)& Very Low (-4.0)\\
        & Proud&{\LARGE 🤩}& Very High (+4.0)& High (+1.0)& Very High (+2.5)\\
        \hline
        \multirow{3}{*}{Fear}& Anxious&{\LARGE 😟}& Low (-1.0)& High (+2.5)& Low (-2.5)\\
        & Afraid&{\LARGE 😨}& Very Low (-5.0)& High (+2.5)& Very Low (-4.0)\\
        & Terrified&{\LARGE 😱}& Very Low (-5.0)& Very High (+4.0)& Very Low (-4.0)\\
        \hline
        \multirow{3}{*}{Anger}& Annoyed&{\LARGE 😒}& Low (-2.5)& Moderate (0.0)& Moderate (-1.0)\\
        & Angry&{\LARGE 😡}& Very Low (-5.0)& High (+2.5)& Moderate (0.0)\\
        & Furious&{\LARGE 🤬}& Very Low (-4.0)& Very High (+4.0)& High (+1.0)\\
        \hline
        \multirow{2}{*}{Sadness}& Sad& {\LARGE 🙁}& Very Low (-4.0)& Low (-2.5)& Very Low (-4.0)\\
        & Devastated&{\LARGE 😭}& Very Low (-4.0)& High (+1.0)& Low (-2.5)\\
        \hline
        \multirow{3}{*}{Shame}& Ashamed&{\LARGE 😳}& Low (-3.0)& Moderate (-1.0)& Very Low (-4.0)\\
        & Embarrassed&{\LARGE 😥}& Very Low (-4.0)& High (+2.5)& Low (-2.5)\\
        & Guilty&{\LARGE 😬}& Very Low (-4.0)& Moderate (0.0)& Very Low (-4.0)\\
    \end{tabular}
    \caption{All 18 emotions used in the study and how they were represented in words, emojis, and Valence-Arousal-Dominance (VAD) scores with values represented lexically and numerically (in parentheses). The category was used to determine which emotions to use but were not integrated into the prompt nor shown to participants.}
    \label{tab:vad_table}
\end{table*}

We prompted LLMs to generate sentences using three words to denote the content of the sentence---which we will refer to as the \textit{keywords}, in addition to the emotion we wanted the sentence to express.

We selected four representations of emotions: 
\begin{itemize}
\item \textit{Words}: English terms for the emotion,
\item \textit{Lexical VAD}: VAD scales expressed in English (Very High, High, Moderate, Low, Very Low),
\item \textit{Numeric VAD}: VAD scales expressed in numeric terms (-5.0 to +5.0), and 
\item \textit{Emojis}\footnote{Emojis were embedded as unicode so that participants would see the set that they were used to seeing on their device, although we recognize this introduces some disparity across participants.}.
\end{itemize}
The numeric values for VAD were generated by normalizing values from \citet{guo_enhancing_2021} (which were in the range of 0.00 to 1.00) to a -5.0 to +5.0 scale, then rounding to the nearest 0.5. 
We also recognize that there will be loss from converting \citeauthor{guo_enhancing_2021}'s scale to a \textit{discrete} numeric representation. However, we found it to lead to easier comprehension.
Before the rounding step, the Numeric VAD values were converted to Lexical VAD. Lexical VAD was mapped to a 5-point scale for simplicity. Due to the rounding step, in some cases there may be an overlap (such as Surprise's +2.5 Arousal and Fear's +2.5 Arousal corresponding to 'Very High' and 'High' respectively). However, this does not pose an issue as Lexical VAD and Numeric VAD are independent representations even though they both capture VAD. 
We hypothesize different strengths from each representation. Lexical VAD may carry extra semantic information from being linguistically derived but Numeric VAD denotes more meaningful separation between points by placing them on an interval scale.

When the LLMs were prompted to generate sentences using the Words representation, we explicitly forbid the models from treating the Emotion as a keyword\footnote{``Do not use the word '\{emotion\}' in the response and express the sentiment in a different way.'' was added to the prompt. The full prompts can be found in Appendix \ref{sec:abs-gpt-prompt}.}, as LLMs have a tendency to ``tell'' instead of ``show'' if not provided enough guidance. This was also not to bias the participants in the Words condition into selecting sentences generated using the Words representation simply because it was using the same word.

We limited each input to three content keywords to give LLMs sufficient context to generate. 
The keywords were sets of arbitrarily-chosen, common words like [\texttt{Place}, \texttt{Great}, \texttt{Korean}], [\texttt{Finals}, \texttt{Semester}, \texttt{Math}].
We generated sentences using GPT-4-Turbo-2024-04-09 \citep{openai_gpt-4_2024} and LLaMA-3.3-70B \citep{grattafiori_llama_2024}, referred to as GPT-4 and LLaMA-3 respectively in this paper. Both models were used with default parameters. GPT-4-Turbo cost less than \$5 in total to generate the sentences. All generations were consistent with the models' indended use.

Due to compute limitations, we used a LLaMA-3 model that was quantized \citep{huang_empirical_2024} using 8-bit weights\footnote{\url{https://huggingface.co/meta-llama/Llama-3.3-70B-Instruct\#use-with-bitsandbytes}}. 
Computing resources for LLaMA-3's generation can be found in Appendix \ref{sec:computing}.
The prompt, used for both models, can be found in Appendix \ref{sec:abs-gpt-prompt}. In the system prompt, we instructed the model to minimize inserting extraneous information while still generating sentences that clearly express an emotion.
Although prior work has shown that LLMs are particularly good at generating sentences from keywords \cite{chen_evaluating_2024}, we did verify this as well (results in Appendix \ref{sec:lexical-constraints}).

For each LLM, we generated a total of 360 sentences, 90 per representation, expressing 18 different emotions. We determined this list of emotions based off \citeauthor{demszky_goemotions_2020} (\citeyear{demszky_goemotions_2020}), ultimately selecting for emotions that were distinct but could also be easily grouped. 
Two sentences were selected randomly for each of the 18 emotions for use in the evaluation, resulting in 36 Representation Alignment questions per condition (i.e., the emotion representation that the participant saw). 
Table \ref{tab:vad_table} shows all the emotions used, their grouping, and their VAD values. 

Overall, 72 sentences for each representation were selected and shown to participants; half for the Representation Alignment questions (Section \ref{sec:repalign}), and the other half for Accuracy and Realism questions (Section \ref{sec:accrealism}). We did this in order to ensure each question had answers from multiple participants.

For generating a sentence with emotions represented as Words or Emojis, we used plain few-shot prompting \citep{brown_language_2020}, selecting exemplars from across the range of positive and negative emotions. For prompting using either form of VAD, we used step-back chain-of-thought prompting \citep{zheng_take_2023}. First, we prompted the model to give an explanation of VAD, then convert it to scale from -5 to +5 if using Numeric VAD, and then gave it few-shot examples. For all representations, we converted the emotions in the exemplars to the representation we intend to generate in.

\section{Evaluation}
\label{sec:Evaluation}

The generated sentences were evaluated via crowdsourcing.
We recruited 100 participants on Prolific for each model for a total of 200 participants, paying them \$14/hr for 15 minutes (\$3.50 each) --- for a total of \$700.
To participate, people were required to agree to the consent form approved by the university's IRB.
Participants needed to be 18+ years of age, living in the United States, and fluent English speakers.
Prolific automatically assigns random identifiers to participants and no identifiable information was collected from participants.
Each participant was randomly assigned one of the four conditions (the four emotion representations) which was the emotion representation that they saw for all questions.

Since we assumed participants would not be used to VAD, people in either VAD condition were shown a short explanation of VAD and given questions with feedback to train them on how to use it (Appendix \ref{sec:vad-training}).
Participants were shown two sets of questions. The first set (Representation Alignment) was used to determine if the representation that the human saw aligned with the LLM's representation. The second set (Accuracy \& Realism) asked the participant to consider how accurately a generated sentence captured a given emotion and how realistic the generated sentence appeared. Both the emotion and sentence from this second set of questions used the same representation. That is, the LLM was given the same representation to generate the sentence as what the user was seeing for the emotion.
We found this setup compelling since the representation that produces the best output from the LLM does not have to perfectly align with the user's internal representation but rather how the user imagines that representation to be expressed. For example, if people believe that Lexical-VAD--prompted LLM output produces sentences that match their interpretation for how a certain Words representation would be expressed, then that is just as valuable information.

A final open-ended question was asked at the end of the survey as an attention check, asking participants to name a piece of media that they have consumed recently and to describe the emotion they felt watching/reading it.
Participants were blocked from copying and pasting text. Any responses that were written in poor English or did not answer the question were removed, and we replaced their data with new responses. We ended up with 26, 25, 28 and 29 participants for Words, Lexical VAD, Numeric VAD and Emojis respectively for GPT-4, and 25 participants for each of the four conditions for LLaMA-3. Due to this difference, we normalized the counts during evaluation and discussion.
The data was analyzed in Python using the packages pandas \& scipy and visualized with matplotlib\footnote{Code for visualizations was generated with help from a mixture of ChatGPT, Claude, and Gemini. All other code was written by hand.}.

\subsection{Human Evaluation 1: Representation Alignment}
\label{sec:repalign}

To evaluate Representation Alignment, we provided an emotion in the representation the person was assigned and showed them sentences generated across all four conditions. The participant was then told to select the sentence that is the best fit for the emotion, as illustrated in Figure \ref{fig:header}. Here, the goal was to determine which representation was most effective at conveying the emotion to the user \textit{as the user understood it} across multiple emotions.

For example, someone in the Words condition would see an emotion (``Anxious'') followed by four sentences. Each of these sentences were generated using a different representation for the emotion ``Anxious''. The sentence order was shuffled before being presented to the participant.
In the example below, 1 was generated using an English Word for the emotion, 2 with Lexical VAD, 3 with Numeric VAD, and 4 with an Emoji.

 \noindent\paragraph{ \textbf{Anxious}}
    \begin{enumerate}
        \item I feel so nervous about my math finals this semester.
        \item I can't believe the semester is almost over, and we've got that big math final coming up soon; it's really time to buckle down and study hard!
        \item I'm really stressed about the math finals this semester.
        \item I'm so happy I passed my math finals this semester!
    \end{enumerate}

Participants were required to answer all questions and were shown 10 of these questions randomly selected from the 36 questions within the participant's condition. The questions were evenly distributed using Qualtrics's ``evenly display questions'' feature. These questions were each answered by 5-9 people (median of 7). Some questions were seen more or less often due to participants not qualifying and the counts not being reset in between participants. 
Full participant instructions can be found in Appendix \ref{sec:app-survey-questions}.

%%%%%%%%

For a user-presented representation $rep_A$ and an LLM-presented representation $rep_B$, 
we defined $rep_A$ as having a strong \textsc{Representation Alignment} with $rep_B$ if
\begin{itemize}
\item participants were most likely to pick $rep_B$ (\textit{selection rate}), and
\item  the average Shannon Entropy for selections $rep_A$ was low, relative to other representations,
\end{itemize}
where $rep_A$ and $rep_B$ may be the same or different representations.
For example, if participants who were shown Lexical VAD emotions were more likely to select sentences generated using Emojis (despite not knowing how the sentence was generated), \textit{and} the entropy for Lexical VAD was relatively low, then Lexical VAD would have good Representation Alignment with Emojis. A random selection rate would be 25.0\%, so any value above this would be notable. 
We will refer to the condition when $rep_A$ = $rep_B$ as \textsc{Self-Alignment}.

We use Shannon Entropy rather than inter-rater agreement because the latter assumes all options are equally likely to be chosen. However, in reality this is unlikely to be the case, and for some emotions or representations, one output may be consistently better than others.

\subsubsection{Human Evaluation 1: Results and Discussion}
\begin{figure}[t]
    \centering
    \includegraphics[width=\linewidth]{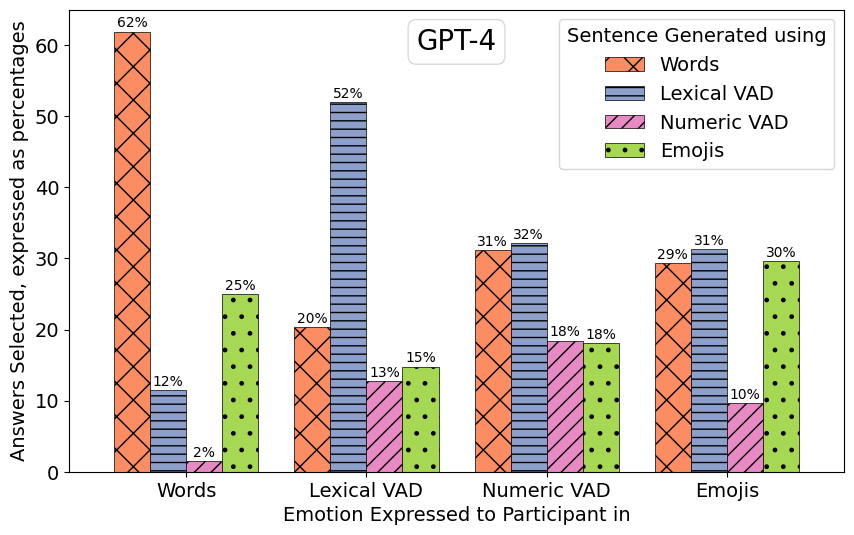}
    \includegraphics[width=\linewidth]{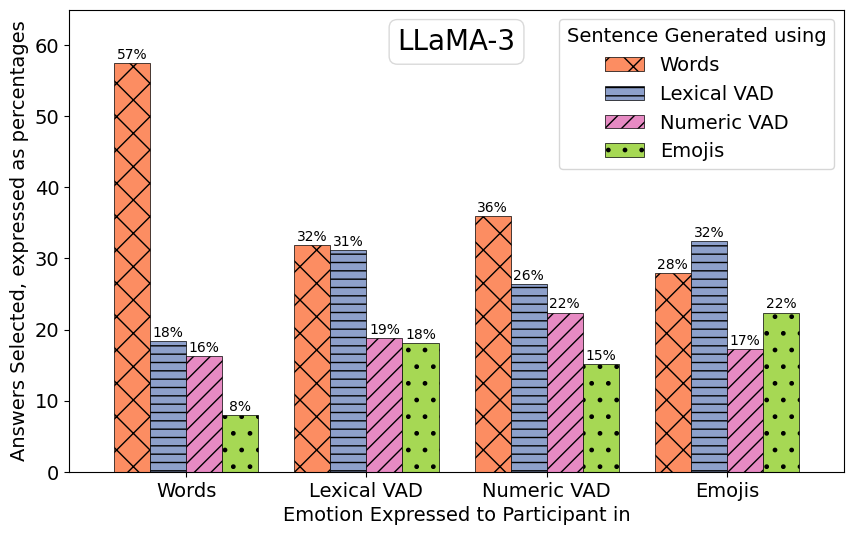}
    \caption{Percentage of times a sentence was selected. Each category on the x-axis corresponds to the condition the participant was in---what representation they saw. The colors delineate what representation was used for sentence generation. Results for GPT-4--generated sentences are on the top, LLaMA-3 on the bottom.}
    \label{fig:emoalign_summary}
\end{figure}

Figure \ref{fig:emoalign_summary} shows the results of the selection rates across representation types and models, while Table \ref{tab:entropy} shows the mean entropy values.
Overall, Words had the best Representation Alignment with Words, regardless of the model. 
This was unsurprising, as humans express emotions in Words most often and LLMs are trained on natural language data. Words had a 61.9\% selection rate using GPT-4 and 57.5\% using LLaMA-3, as well as Shannon Entropies of 0.32 and 0.42 respectively.

\begin{table}[b]
    \centering
    \begin{tabular}{l|c|c}
       \textbf{Participant's}  & \multicolumn{2}{c}{\textbf{Entropy}$\downarrow$}\\
         \textbf{Representation}&  \textbf{GPT-4}& \textbf{LLaMA-3} \\\hline
         Words&  \textbf{.32} & \textbf{.42}\\
         Lexical VAD&  \underline{.61} & .72\\
         Numeric VAD&  .70 & .63\\
         Emojis&  .67 & \underline{.52}\\
    \end{tabular}
    \caption{Shannon Entropy values showing the amount of variability in responses for each representation that was presented to participants. 
Bolded values show the most agreement, underlined are second most.}
    \label{tab:entropy}
\end{table}

More surprisingly, Lexical VAD also has a high self-selection rate of 52.0\% for GPT-4, with the second lowest entropy value at 0.61. 
While the agreement is worse than that of Words, the high selection rate is noteworthy. This occurs even though the participants and the LLMs are given different instructions and prompts. (Appendix \ref{sec:abs-gpt-prompt} and \ref{sec:vad-training}) This indicates both humans and LLMs may be drawing on similar ideas or memorized information when considering the emotion representation. On the other hand, 
LLaMA-3's output never matched people's expectations of Lexical VAD, regardless of what the model was prompted with, resulting in a higher entropy value.

\begin{figure*}[h]
    \centering

    \includegraphics[width=0.9\linewidth]{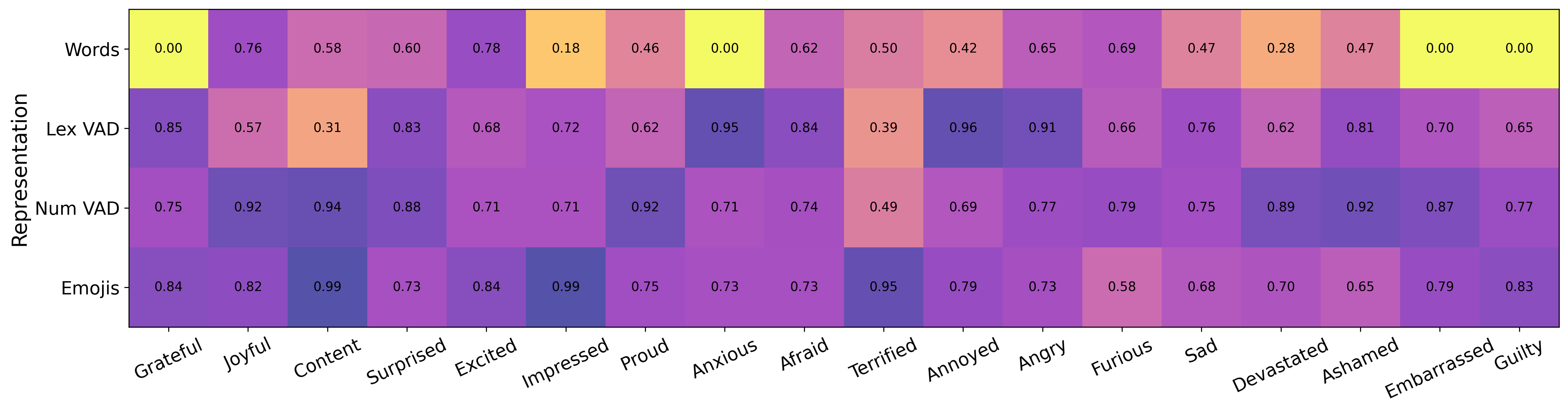}

    \includegraphics[width=0.9\linewidth]{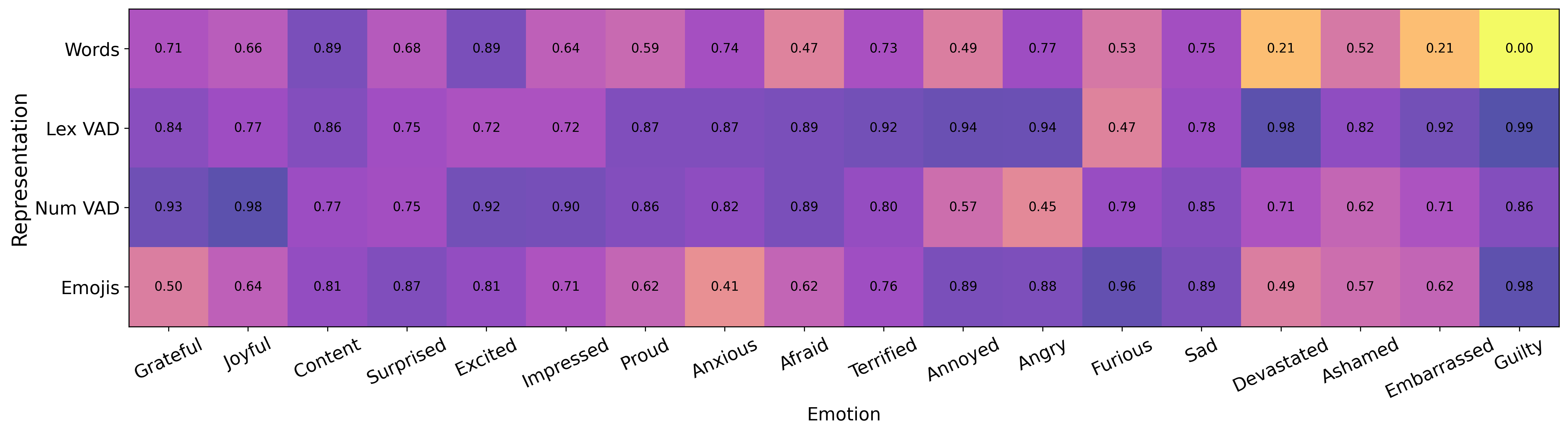}
    
    \caption{Heat maps for Shannon entropy of each emotion across representations. Lower (brighter) values are better, denoting more ``agreement'' between participants and the LLM. Top: GPT-4, bottom: LLaMA-3.}
    \label{fig:Shannon_Entropy_Heatmap}
\end{figure*}

Numeric VAD, by contrast, had poor alignment, with the worst entropy values for both LLMs. We interpret this to mean that people could not consistently agree on an expected output for Numeric VAD emotions. One possibility is that participants struggled to conceptualize the numbers but were able to understand them more easily when using discrete words (leading to Lexical VAD doing better). This is prudential given some recent works offer users control over generative models using VAD \citep{tang_emomix_2023, he_emoticrafter_2025}.

Emojis, which have been shown to be subjective and ambiguous \citep{miller_understanding_2017}, did not show particularly good alignment. However, Emojis did have some Representation Alignment with Lexical VAD for LLaMA-3 --- 32.40\% selection rate and .52 entropy. The trend for GPT-4 was less pronounced, with a selection rate of 31.38\% and .67 entropy. 
Because there was a mismatch in representations it hints at that people's mental model of emojis may be similar to how LLaMA-3 works with Lexical VAD. 

One potential explanation is that Lexical VAD for LLMs and Emojis for humans capture the same amount of information for imprecise emotions. In other words, LLMs can easily decompose Lexical VAD into emotions into components while Emojis are discrete symbols to text-based models like LLaMA-3. Meanwhile, humans can assign meaning to individual facial features in Emojis but may struggle to do the same with VAD scales. Further research will need to be done to verify this claim.  

Ultimately, the highest Representation Alignment was Self-Alignment with Words and, for GPT-4, Lexical VAD Self-Alignment was a close second.

We also broke down Shannon Entropy values by emotion (see Figure \ref{fig:Shannon_Entropy_Heatmap}).
For individual emotions per representation, we found the results are fairly similar, with most values between 0.70 to 0.95 meaning poor agreement, with some outliers. 
The small gap in the models' agreement with participants for certain emotions point to potential differences in training.  It is unclear why, for example, GPT-4 is much better at using Words to generate ``grateful'' sentences than LLaMA-3. To have emotions that people agree or disagree with for both models could mean a variety of things. For instance, is it that people agree more about whether a sentence is expressing ``guilty'', is there more ``guilty'' data both LLMs are trained, or does something inherent to the transformer architecture (such as attention) lend itself to picking up on ``guilty'' text better? A study on model architecture, training data, and conceptual alignment should be run to see if this trend continues.

%%%%%%%%

\subsection{Human Evaluation 2: Accuracy \& Realism}
\label{sec:accrealism}

\begin{figure}[t]
    \includegraphics[width=1\linewidth]{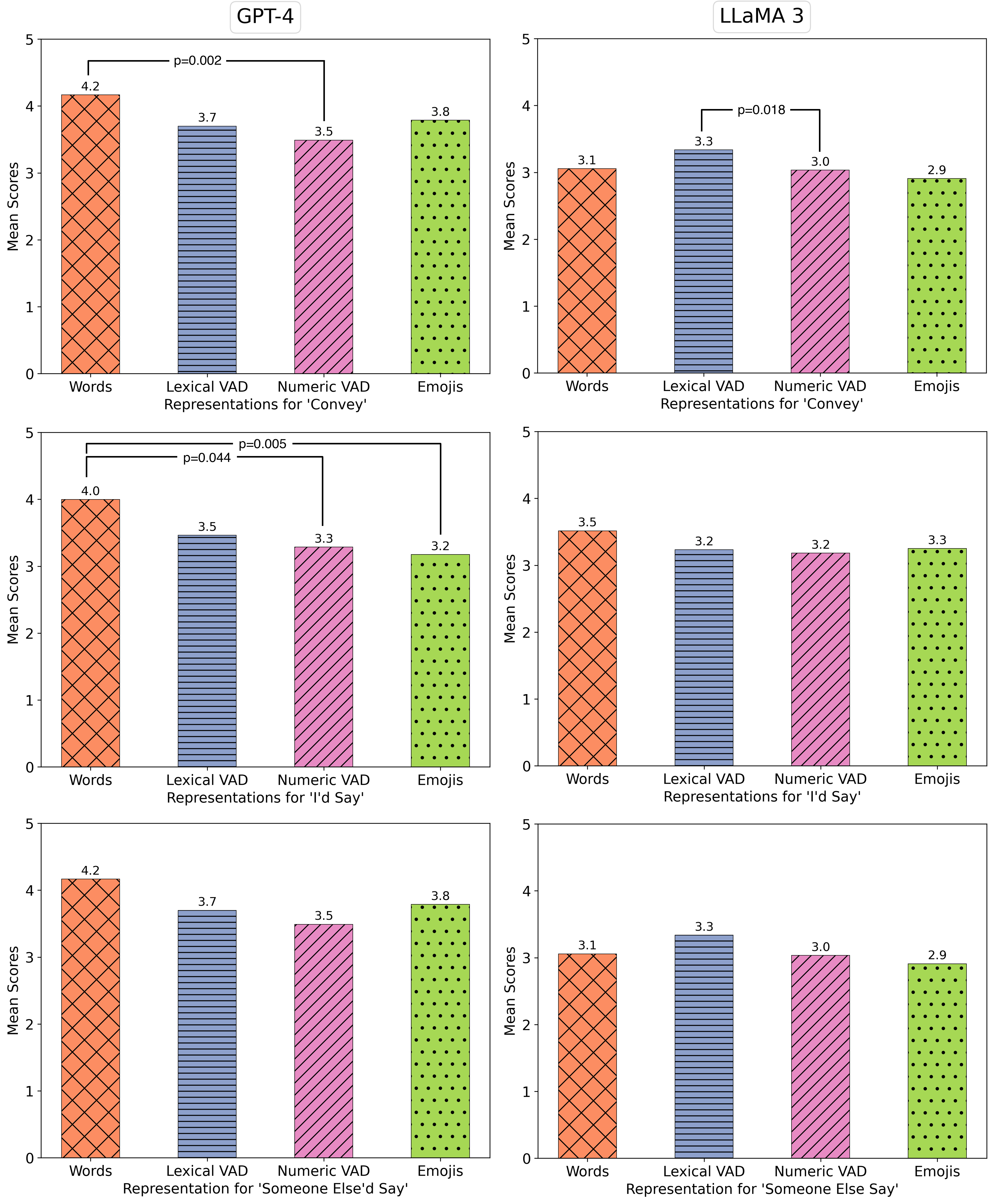}
   \caption{Left: GPT-4, Right: LLaMA-3. In order from Left to Right and Top to Bottom:\\ a, b. Histograms of the Mean Scores for `Convey'\\
    c, d. Histograms of the Mean Scores for `You'd say'\\ e, f. Histograms of the Mean Scores for `Someone Else'd Say'}
   \label{fig:gpt4_realism_summary}
\end{figure}
\begin{figure*}[t]
    \centering
    \includegraphics[width=0.9\linewidth]{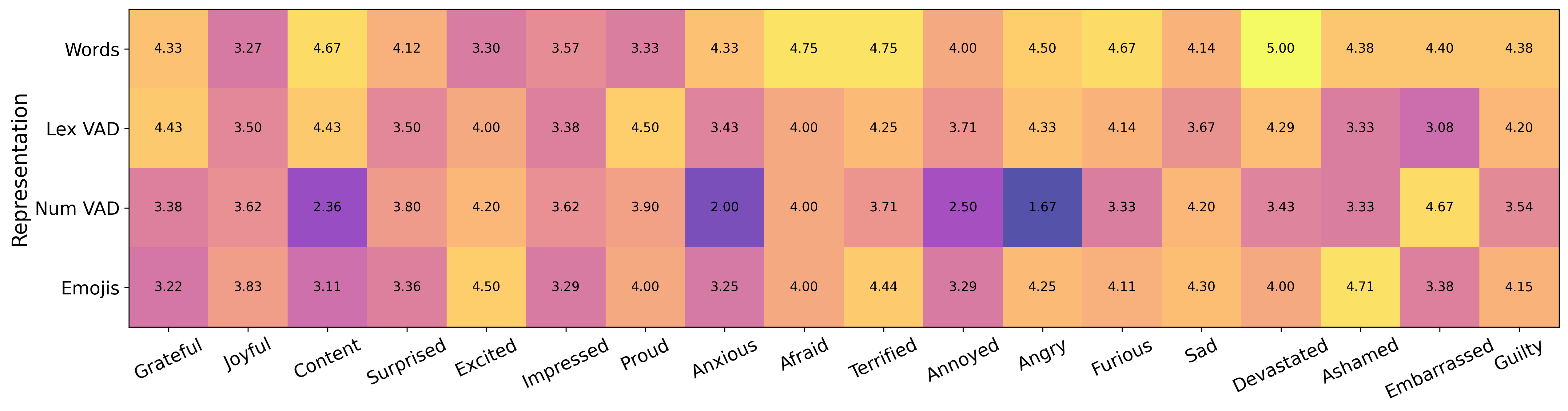}
    \includegraphics[width=0.9\linewidth]{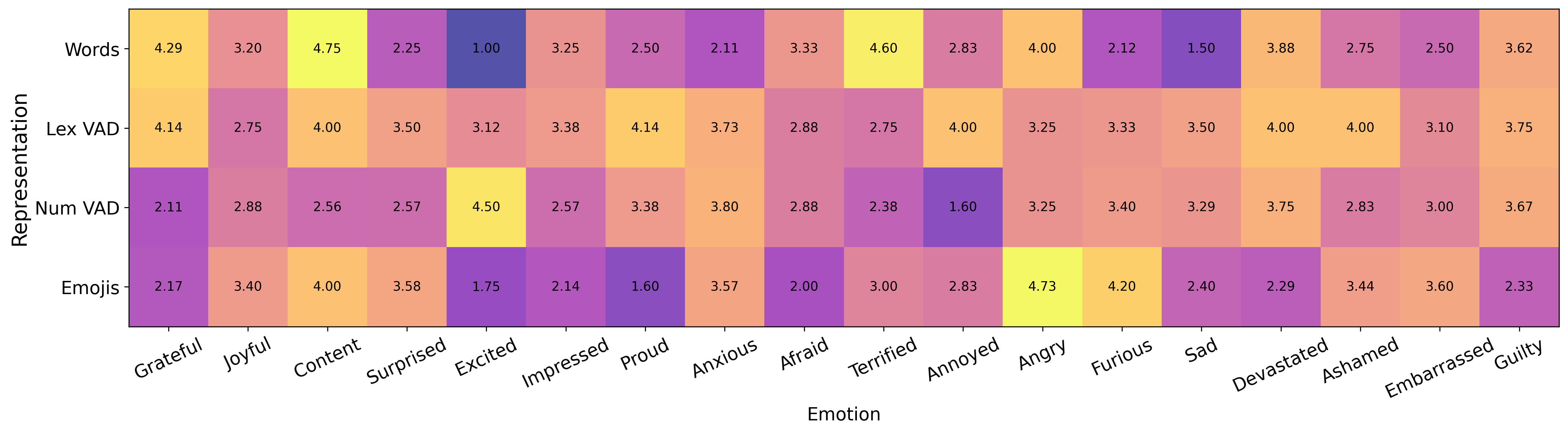}
    \caption{Mean ``Convey'' Scores for each emotion per representation. Higher (brighter) values are better. The top map shows results for GPT-4, while the bottom map is LLaMA-3.}
    \label{fig:Mean_Convey_Heatmap}
\end{figure*}

The second set of questions asked the participant to consider how accurate or realistic a pair of an emotion and a sentence generated with that emotion is. The participants were asked to rate the three questions (shown in the example below) on a 5-point Likert slider, using the labels Not at all (1), Slightly (2), Moderately (3), Very (4), Extremely (5).  For example:

\begin{quote}
    For the following questions, consider the emotion represented by these VAD values: \textbf{Very High Valence, Moderate Arousal, Low Dominance}

    And this sentence: \textbf{``This place has great Korean food; it always makes me so happy!''}
\newpage
  How much does the sentence...\\
    - Convey the emotion above? \\
    - Sound like something that you would say?\\
    - Sound like something that someone else would say?
\end{quote}

We will refer to these three questions as ``Convey'', ``You'd say'', and ``Someone Else'd say'', respectively.
The ``Convey'' question allows us to assess how well the generated sentence accurately expresses the emotion. The ``You'd say'' and ``Someone Else'd say'' questions are to measure how realistic and human-like the sentence is. The slider's default value for each question was Moderately (3). Participants were required to answer all questions.

Each user was given 6 questions of this type, randomly selected from the 36 total accuracy \& realism questions (emotion-sentence pairings). Each question was answered by 3.7 people on average (1-8 participants each, median of 4), with 3 questions not answered and therefore not considered in the analysis.

\subsubsection{Human Evaluation 2: Results and Discussion}

The average Likert scores for ``Convey'', ``You'd say'', and ``Someone Else'd say'' across conditions (the emotion representations) and for both models can be found in Figure \ref{fig:gpt4_realism_summary}.
We ran an ANOVA statistical significance test on the Likert ratings for the three questions for both GPT-4 and LLaMA-3.

The ANOVA showed that, for GPT-4, the emotion representation had a statistically significant effect on the rating for ``Convey'' and ``I'd say'', both $p < .01$.
For LLaMA-3, ANOVA showed the emotion representation had a statistically significant effect on the rating for ``Convey'' and ``Someone Else'd say'', both $p < .05$.
A pairwise t-test was run on the statistically significant results, which showed that Words was significantly better at ``Convey'' than Numeric VAD for GPT-4 ($p=0.002$) and Lexical VAD was also significantly better at ``Convey'' than Numeric VAD for LLaMA-3 ($p=0.018$) -- further showing that Numeric VAD scores under-perform.

For ``You'd say'' questions, Words is significantly better than both Emojis ($p=0.005$) and Numeric VAD ($p=0.044$) when using GPT-4 to generate plausible-sounding sentences. This shows that Words would most likely be the preferred representation to capture an emotion appropriately.
Additionally, LLaMA-3's generated sentences are overall perceived as slightly worse for conveying the emotion and sounding realistic. However, this is in line with the relative performances of each model.

We show mean scores for the ``Convey'' question broken down by emotion in Figure \ref{fig:Mean_Convey_Heatmap}. The scores are relatively high across the board, with some outliers. For instance, GPT-4 using Words seemingly struggles most with sounding realistically ``excited'' or ``proud'', while using Numeric VAD makes GPT-4 struggle with emotions like ``anxious'' and ``angry''. LLaMA-3, on the other hand, sounds unrealistic when using ``sad'' or ``excited'' as Words.

Despite guardrails that make LLMs more amicable \citep{sharma_towards_2023, zhou_is_2024}, we show that models can struggle to generate certain positive emotions under complex conditions, and that different emotions are captured best by different representations. Out of the representations we looked at, Words and Lexical VAD were the best for producing realistic sentences.

\section{Conclusion}

We believe that research studying representations should measure the way they are perceived and represented to both humans and LLMs. 
This work introduced the evaluation paradigm of \textsc{Representation Alignment} for determining if humans perceive representations in the same way that LLMs use them.
Effectively evaluating how well human expectations for conceptual knowledge align with LLM output is a largely understudied problem, especially with downstream contexts in mind such as AAC.
We adopted the problem of emotional sentence generation based on keywords and used Representation Alignment to study how various emotional representations align with participants' expectations. We hope that this experimental setup will be used to study other types of conceptual alignment as well, such as value alignment.

Using this paradigm, we showed that emotion-based Words provided the strongest alignment between an LLMs' understanding of an emotion \& how it generates with it and humans' expectations of its use. Lexical VAD is a close second for both representation alignment and for accuracy and realism. 
In AAC applications, this can enable users to select their intended emotion or tone more precisely.

\section{Limitations}

Our study uses GPT-4-Turbo-2024-04-09 and LLaMA-3.3-70B with 8-bit weights due to compute constraints. It is possible that a more recent or bigger models such as GPT-4o or LLaMA-3-3.1-405B would give different results. Due to funding limitations, this study did not look at every state-of-the-art model available. However, we hope that this can serve as a proof-of-concept for other researchers who are interested in using our methodology. 

This study was limited to English only. Other languages may give different outcomes. The participants were required to be native English speakers, which allowed us to control the quality of responses. However, it makes the results less generalizable in a global context. Our participants are also sampled from the general public and not necessarily AAC users. Assistive technologies need to be specialized to fit a person's needs, and therefore it is not enough to find overall Representation Alignment but also Representation Alignment for specific users. We believe that Representation Alignment will contribute to ease of use, but individual preferences and needs also matter.

Additionally, while we offer some training for interpreting VAD (whether lexical or numeric), it is still difficult to grasp without any prior knowledge and the heavy cognitive load needed to ``calculate'' these emotions may have affected the results.

Furthermore, while the emojis were selected based on what we believe was the best fit for each English word, there could be cultural discrepancies over what each emoji might mean. For example, different online communities might use {\Large 😥} to mean embarrassed, worried, or relieved, depending on the context they see it used in and how their system displays it. More in-depth analyses of emoji use should gather general---like \cite{Warriner_2013} have done with Numeric VAD---and/or community-specific use to determine the exact meaning.

\section{Ethical Considerations}

LLMs can be used in many helpful cases, but they can also be used to impersonate individuals, or generate toxic or deceptive texts. We acknowledge that improving controlled text generation capabilities may also make it easier to generate the above, and that strong guardrails are necessary. In this work, we study how LLMs align with human emotions in terms of representations. This does not indicate that LLMs themselves are capable of possessing emotions, only that they are capable of recognizing and generating emotion when encoded in some textual form to the extent their model design and training data allows.

Having LLMs generate text for people can lead to questions of authorship. These issues are exacerbated when LLMs are introduce into AAC applications, potentially leading people to question the agency of AAC users and the validity of what they say. LLMs need to be carefully implemented into AAC applications with clear guidelines on how the AAC user should interface with the AI and provide potential scenarios where LLM use may not be useful or helpful. Additionally, many LLMs are too big to put onto tablets or phones and therefore require a network connection, which adds security risk.

 \section*{Acknowledgments}
 We would like to thank all of our participants for their time and effort. We would also like to thank Foad Hamidi and others who have given us feedback throughout the study, and to the UMBC High Performance Computing Facility for helping us run our LLaMA experiments. This research is partially supported by a 2024 UMBC COEIT interdisciplinary proposal (CIP) Award.

\bibliographystyle{acl}
\bibliography{anthology,custom}

\newpage

\appendix

\section{Computing Information}
\label{sec:computing}
LLaMA-3 was run on UMBC's high performance computing cluster called Ada, which at the time, had:
\begin{itemize}
\item 4 8x RTX 2080 Ti GPUs with 384 GB CPU memory \& 11GB GPU memory each, 
\item 7 8x RTX 6000 GPUs with 384 GB CPU memory \& 24GB GPU memory each, and
\item 2 8x RTX 8000 GPUs with 768 GB CPU memory \& 48GB GPU memory each.
\end{itemize}
Each node had 48 threads and two 24-core Intel Cascade Lake CPUs.
The node that was assigned for a particular job was random.

\section{Prompts}
\label{sec:abs-gpt-prompt}

\textbf{System Prompt}:

\begin{tcolorbox}
"You are engaging in a conversation with a human. Respond to the following line of dialogue based on the given emotion and the following keywords. Just add connective words and do not add any new information to the output sentence. Do not use the word '{emotion}' in the response and express the sentiment in a different way.
\end{tcolorbox}

The last line is only for when prompting with Words. 
\subsection{Words}

\begin{tcolorbox}[colback=orange!25!]

Here are some examples:\\

Emotion: Proud\\
Keywords: 'running', 'marathon', 'first'\\
Dialogue: Running my first marathon felt like such a huge accomplishment!\\
    
Emotion: Sad\\
Keywords: 'banana', 'plant', 'brown'\\
Dialogue: It really sucks that my banana plant's leaf is turning brown\\
    
Now, respond to the following. Remember, do not use the word \{emo\_\} in the dialogue:\\
    
Emotion: \{emo\_\}\\
Keywords: \{kwds\_\}\\
Dialogue:
\end{tcolorbox}

\subsection{Lexical VAD}

\begin{tcolorbox}[enhanced, breakable, colback=blue!15!]
Valence refers to the intrinsic attractiveness or averseness of an event, object, or situation. In the context of emotions in text, valence represents the positivity or negativity of the emotion expressed. For example, words like "happy," "joyful," or "excited" have positive valence, whereas words like "sad," "angry," or "frustrated" have negative valence.\\
    
It essentially measures the degree of pleasantness or unpleasantness of the emotion.\\

Arousal indicates the level of alertness, excitement, or energy associated with an emotion. It ranges from high arousal (e.g., excitement, anger) to low arousal (e.g., calm, boredom). In text, high-arousal words might include "thrilled," "furious," or "ecstatic," while low-arousal words could be "relaxed," "content," or "lethargic."\\

This dimension measures how stimulating or soothing the emotional state is.\\
        
Dominance reflects the degree of control, influence, or power that one feels in a particular emotional state. High dominance implies feelings of control and empowerment, while low dominance suggests feelings of submissiveness or lack of control. In text, emotions like "confident," "powerful," or "authoritative" would have high dominance, whereas "helpless," "weak," or "submissive" would have low dominance.\\
    
It gauges the extent to which an individual feels in control or overpowered by the emotion.\\
        
Now, assume you are a normal human. Say a line of natural dialogue based on the given keywords. Just add connective words and do not add any new information to the output sentence.\\
        
For example: \\
    
Emotion: Very High Valence, High Arousal, Very High Dominance\\
Keywords: 'running', 'marathon', 'first'\\
Dialogue: Running my first marathon felt like such a huge accomplishment!\\
    
Emotion: Very Low Valence, Low Arousal, Low Dominance\\
Keywords: 'banana', 'plant', 'brown'\\
Dialogue: It really sucks that my banana plant is turning brown\\
            
Emotion: Very High Valence, Very High Arousal, High Dominance\\
Keywords: "visit", "parents", "month"\\
Dialogue: I'm finally going to visit my parents next month!\\
    
Now, respond to the following:\\
Emotion: \{v\_\}, \{a\_\}, and \{d\_\}.\\
Keywords: \{kwds\_\}\\
Dialogue:
\end{tcolorbox}

\subsection{Numeric VAD}

\begin{tcolorbox}[enhanced, breakable, colback=purple!25!]
Valence refers to the intrinsic attractiveness or averseness of an event, object, or situation. In the context of emotions in text, valence represents the positivity or negativity of the emotion expressed. For example, words like "happy," "joyful," or "excited" have positive valence, whereas words like "sad," "angry," or "frustrated" have negative valence.\\

It essentially measures the degree of pleasantness or unpleasantness of the emotion.\\              
Arousal indicates the level of alertness, excitement, or energy associated with an emotion. It ranges from high arousal (e.g., excitement, anger) to low arousal (e.g., calm, boredom). In text, high-arousal words might include "thrilled," "furious," or "ecstatic," while low-arousal words could be "relaxed," "content," or "lethargic."\\
    
This dimension measures how stimulating or soothing the emotional state is.\\
        
Dominance reflects the degree of control, influence, or power that one feels in a particular emotional state. High dominance implies feelings of control and empowerment, while low dominance suggests feelings of submissiveness or lack of control. In text, emotions like "confident," "powerful," or "authoritative" would have high dominance, whereas "helpless," "weak," or "submissive" would have low dominance.\\
    
It gauges the extent to which an individual feels in control or overpowered by the emotion.\\

Here's how each dimension can be defined on a scale from -5.0 to 5.0:\\
    
Valence:\\
    -5.0: Extremely negative (e.g., intense sadness, extreme anger)\\
    -2.5: Moderately negative (e.g., mild annoyance, slight disappointment)\\
    0.0: Neutral (e.g., indifferent, no strong emotional reaction)\\
    2.5: Moderately positive (e.g., mild pleasure, slight happiness)\\
    5.0: Extremely positive (e.g., intense joy, deep love)\\
    
Arousal:\\
    -5.0: Extremely low arousal (e.g., deep sleep, total relaxation)\\
    -2.5: Moderately low arousal (e.g., relaxed, slightly tired)\\
    0.0: Neutral arousal (e.g., alert but not excited, calm)\\
    2.5: Moderately high arousal (e.g., interested, mildly excited)\\
    5.0: Extremely high arousal (e.g., highly excited, very agitated)\\
    
Dominance:
    -5.0: Extremely low dominance (e.g., feeling completely powerless, totally submissive)\\
    -2.5: Moderately low dominance (e.g., somewhat submissive, slightly dominated)\\
    0.0: Neutral dominance (e.g., feeling neither in control nor dominated)\\
    2.5: Moderately high dominance (e.g., feeling somewhat in control, slightly assertive)\\
    5.0: Extremely high dominance (e.g., feeling very powerful, completely in control)\\
                    
These scales provide a way to quantify and compare the emotional dimensions in a structured manner.\\
        
Now, assume you are a normal human. Say a line of natural dialogue based on the given keywords. Just add connective words and do not add any new information to the output sentence.\\
        
For example:\\
    
Emotion: Valence: 4.0, Arousal: 1.0, Dominance: 2.5\\
Keywords: 'running', 'marathon', 'first'\\
Dialogue: Running my first marathon felt like such a huge accomplishment!\\
    
Emotion: Valence: -4.0, Arousal: -2.5, Dominance: -4.0\\
Keywords: 'banana', 'plant', 'brown'\\
Dialogue: It really sucks that my banana plant is turning brown\\
            
Emotion: Valence: 2.5, Arousal: 4.0, Dominance: 1.0\\
Keywords: "visit", "parents", "month"\\
Dialogue: I'm finally going to visit my parents next month!\\
    
Now, respond to the following:\\
Emotion: \{v\_\}, \{a\_\}, and \{d\_\}.\\
Keywords: \{kwds\_\}\\
Dialogue:
\end{tcolorbox}

\subsection{Emojis}

\begin{tcolorbox}[enhanced, breakable, colback=green!10!]
You are engaging in a conversation with a human. Respond to the following line of dialogue based on the given emotion and the following keywords.\\

Just add connective words and do not add any new information to the output sentence. The response should be exactly one line with nothing else other than the responding dialogue.\\
    
For example: \\
    
Emotion: {\Large 🤩}\\
Keywords: 'running', 'marathon', 'first'\\
Dialogue: Running my first marathon felt like such a huge accomplishment!\\
    
Emotion: {\Large 🙁}\\
Keywords: 'banana', 'plant', 'brown'\\
Dialogue: It really sucks that my banana plant is turning brown\\
        
Emotion: {\Large 😃}\\
Keywords: "visit", "parents", "month"\\
Dialogue: I'm finally going to visit my parents next month!\\
    
Now, respond to the following:\\
Emotion: \{emo\_\}\\
Keywords: \{kwds\_\}\\
Dialogue:
\end{tcolorbox}

\section{VAD Training}
\label{sec:vad-training}

In the following sections, we give the training instructions given to the participants verbatim. Figures \ref{fig:lexical_vad_train} and \ref{fig:numeric_vad_train} were shown to the participants at the time of training, but not when they were answering the survey questions.

\subsection{Lexical VAD}

For this study we will be using a popular model used for quantifying emotion called the Valence-Arousal-Dominance (VAD) model.

1. Valence — How pleasant you feel. A low valence would mean you are feeling negative/unpleasant whereas high valence would mean you are feeling positive or pleasant.

2. Arousal — How engaged or alert you feel. Low arousal would mean that you are more on the calmer or sleepier extreme, while high arousal would mean you are more active and energetic.

3. Dominance — How much control you have over what you feel. Low dominance implies no control and High dominance implies feeling very much in control of your emotion.

\begin{figure}
    \centering
    \includegraphics[width=1\linewidth]{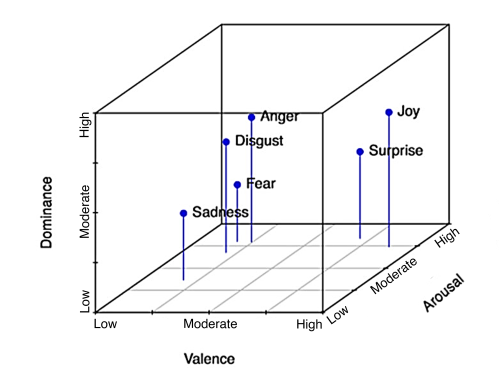}
    \caption{Lexical VAD visualization}
    \label{fig:lexical_vad_train}
\end{figure}

Please take a moment to study the figure [\ref{fig:lexical_vad_train}], as it might be helpful for visualizing the above. It shows some common emotions we usually feel and how they map to the VAD model.

For example, consider the difference between Angry and Furious. Both of these would have low valence and moderate-to-high dominance. Being Angry has high arousal as it takes a lot of energy to feel so. Being Furious would take even more energy, as you might feel like you're about to burst. So, Angry would have High Arousal while Furious would have Very High Arousal.

Similarly, consider the difference between feeling Grateful and Joyous. Both of them are positive emotions. Grateful should have High Valence, as you are feeling pleased but not over the top, while Joyous will have Very High valence as you are really happy and elated.

Before you begin, you will go through a series of questions designed to help you understand the VAD model of emotion, followed by a practice question.

\noindent\rule{\linewidth}{1pt}\\

[We include a sample of the questions here:]

\begin{quote}
    What is the emotion that corresponds to VAD values High Valence, Very High Arousal and Moderate Dominance?

    Surprise\\
    Joy\\
    Anger\\
\end{quote}

\begin{quote}
    Correct Answer: Surprise.
    
    High Valence indicates this is more of a positive emotion. Very High Arousal means there is a lot of energy behind this, while Moderate Dominance shows that you are not entirely in control. This could be either Joy or Surprise, but having higher arousal and lower dominance suggests Surprise is the answer.
\end{quote}

\subsection{Numeric VAD}

For this study we will be using a popular model used for quantifying emotion called the Valence-Arousal-Dominance (VAD) model. In this model, the X, Y, and Z axes span only from -5 to 5 and can be defined as follows

1. Valence — How pleasant you feel on a range from -5 to 5. Here, -5 would mean you are feeling very negative/unpleasant whereas a 5 would mean you are feeling very positive or pleasant.

2. Arousal — How engaged or alert you feel on a range from -5 to 5. -5 would mean that you are more on the calmer or sleepier extreme while 5 would mean you are more active and energetic.

3. Dominance — How much control you have over what you feel on a range from -5 to 5. In this case, -5 implies no control and 5 implies feeling very much in control of your emotion.

\begin{figure}
    \centering
    \includegraphics[width=1\linewidth]{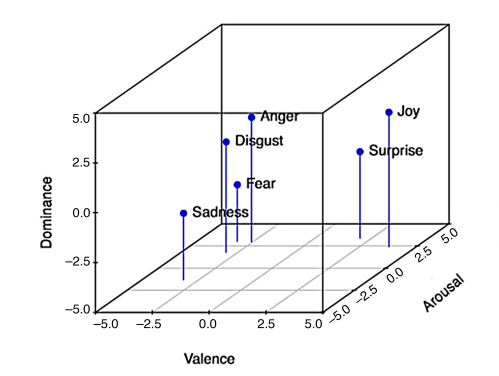}
    \caption{Numeric VAD Visualization}
    \label{fig:numeric_vad_train}
\end{figure}

Please view the figure [\ref{fig:numeric_vad_train}] for a visual representation of these ranges.

Before you begin, you will go through a series of questions designed to help you understand the VAD model of emotion. In the first set of questions you will be provided the numerical values and need to choose the discrete emotion those VAD values correspond to. Then you will be given a practice question similar to the rest of the questions in the survey.

\noindent\rule{\linewidth}{1pt}\\

[We include a sample of the questions here:]

\begin{quote}
    What is the discrete emotion that corresponds to VAD values -2.5 (Valence), 2.5 (Arousal), and 2.0 (Dominance)?

    Sadness\\
    Anger\\
    Fear
\end{quote}

\begin{quote}
    Anger is the correct answer!

    -2.5 Valence indicates this is a negative or unpleasant emotion. 2.5 arousal means it takes a lot of energy to feel this way. Therefore, it cannot be Sadness. 2.0 Dominance means you are somewhat in control of how you feel, so it is unlikely to be Fear either. So, Anger is the most appropriate option.
\end{quote}

\section{Survey Questions}
\label{sec:app-survey-questions}

[Any text appearing within brackets like this in the following section is a note and did not appear in the survey.]\\

\textbf{Informed Consent Information}

\begin{quote}
\textbf{Informed consent:}
You must be 18 years or older to participate in this study.

The purpose of this study is to see if large language models like ChatGPT describe emotions in the same way that people do. You are being asked to volunteer because you are a native English speaker.

You will be shown a series of 4 different sentences and need to determine if each sentence conveys a certain emotion. Note that the emotion may be displayed in an abstract way. You will be taught how to read this abstraction before answering the questions.

The survey may take about 15 minutes to complete.

You are welcome to withdraw or discontinue participation at anytime, but due to the volume of participants expected from crowdsourcing, we will not be paying participants for incomplete surveys. If you withdraw from the study or do not complete the survey, your data will be deleted.

Please take your time and do the best you can. There are no right or wrong answers, but we reserve the right to not pay if we determine that you are not following directions or taking the task seriously.

All data obtained will be anonymous. There is no way for us to find out who you are, and your data will not be shared with any other parties under any circumstance.

Any information learned and collected from this study in which you might be identified will remain confidential. The investigator will attempt to keep your personal information confidential. To help protect your confidentiality, your data will only be linked to a randomly-assigned ID. Any information required to pay you (i.e., username) will be kept in a spreadsheet on a secure server separate from the other data you provide.

Only the investigator and members of the research team will have access to these records. If information learned from this study is published, you will not be identified by name and all results will be reported in aggregate. By signing this form, however, you allow the research study investigator to make your records available to the University of Maryland, Baltimore County's Institutional Review Board (IRB) and regulatory agencies as required to do so by law.
\end{quote}

\noindent\rule{\linewidth}{1pt}\\

\textbf{Introduction to the questions}

\begin{quote}
In the following survey, you will be asked questions based on understanding and recognizing emotions in text. Following the practice questions, there will be 16 questions in total.

The emotions will be described as a word.

For example: Angry, Happy, Annoyed
\end{quote}

\noindent\rule{\linewidth}{1pt}

\begin{quote}
In the following survey, you will be asked questions based on understanding and recognizing emotions in text. Following the practice questions, there will be 16 survey questions in total.

The emotions will be described in terms of valence, arousal and dominance. For example: High Valence, High Arousal, Low Dominance.

In the next page, we will explain what these terms are and how they relate to emotions.
\end{quote}

\noindent\rule{\linewidth}{1pt}

\begin{quote}
In the following survey, you will be asked questions based on understanding and recognizing emotions in text. Following the practice questions, there will be 16 survey questions in total.

The emotions will be described in terms of valence, arousal and dominance. For example:
Valence: -2.0, Arousal: 3.0, Dominance: 4.0.

In the next page, we will explain what these terms are and how they relate to emotions.
\end{quote}

\noindent\rule{\linewidth}{1pt}

\begin{quote}
In the following survey, you will be asked questions based on understanding and recognizing emotions in text. Following the practice questions, there will be 16 survey questions in total.

The emotions will be described using emojis.

For example: {\Large 😄}, {\Large 😯}, {\Large 🤬}
\end{quote}

\noindent\rule{\linewidth}{1pt}\\

\textbf{Representational Alignment Instructions (same for all representations)}

\begin{quote}
For each question below, you will be shown an emotion and a set of sentences. Given the specified emotion, pick the sentence that is the best fit.

Note: In some cases, one or more of the choices might be identical. If you feel that sentence is the best fit, feel free to pick any one. Also, the sentences are not meant to be ironic or sarcastic.
\end{quote}

\noindent\rule{\linewidth}{1pt}\\

\textbf{Accuracy and Realism Instructions (same for all representations)}

\begin{quote}
For the next set of questions, you will be given a sentence and an emotion described in a word.

We will ask you to rate the sentence based on how well it conveys the given emotion, and how realistic it sounds (i.e. it sounds like something a person would say). Please rate how well the sentence reflects each statement.
\end{quote}

\noindent\rule{\linewidth}{1pt}\\

[Bonus Question --- we used this as a filter to gauge how much attention users gave to the survey. In a handful of cases, we removed answers to this question that seemed to be written in extremely poor English or written by a language model.]

\begin{quote}
Think of a movie, television show, or book that you watched or read recently that made you feel a strong emotion.

Please share the name of the movie, show, or book. Then tell us what that emotion was in plain English, and why did you feel that way?\\
(Your response should be at least 30 characters long.)
\end{quote}

 \section{Additional Analysis: Adherence to Keywords}
 \label{sec:lexical-constraints}
 \begin{table}[ht]
     \centering
     
     \begin{tabular}{l|c|c|c|c}
          \textbf{Model} &  \textbf{One} &  \textbf{Two} & \textbf{Three}  &  \textbf{Acc} \\\hline
          GPT-4, 1x&  1.00&  1.00&  .936& .978\\
          LLaMA-3, 1x & .908& .897& .781&.862\\
          LLaMA-3, 3x & .969& .969&  .850&.930\\
          LLaMA-3, 10x & .981& .981&  .853&.938\\
     \end{tabular}
     \caption{Percentage of generated sentences that contain the respective number of keywords. *x indicates the \# of times the model was sampled. Accuracy is percentage of the three keywords that are correct, averaged across all sentences. }
\label{tab:lexical_coverage}
 \end{table}

 Since we were constraining the model to generate using content-based keywords verbatim, we lemmatized each inputted keyword and compared them against the words in the generated sentence. Shown in Table \ref{tab:lexical_coverage},
 all sentences generated by GPT-4 had at least two keywords present, and 93.6\% had all three.
 By contrast, 9.17\% of LLaMA-3's generated sentences had no keywords present at all. Re-sampling improved results, reducing this percentage to just 1.94\%.
 Similar to \citet{chen_evaluating_2024}, this overall trend shows that conditioning on emotions in addition to content keywords does not significantly impair the LLM's ability to copy the keywords when generating.

\end{document}